\documentclass{article}

\usepackage[preprint]{neurips_2024}

\usepackage[utf8]{inputenc} 
\usepackage[T1]{fontenc}    
\usepackage{hyperref}       
\usepackage{url}            
\usepackage{booktabs}       
\usepackage{amsfonts}       
\usepackage{nicefrac}       
\usepackage{microtype}      
\usepackage{xcolor}         
\usepackage[most]{tcolorbox}
\usepackage{xparse}
\usepackage{textcomp} 
\usepackage{wrapfig}
\usepackage{fontspec}   
\usepackage{fontenc}
\usepackage[utf8]{inputenc}

\title{Instruction Mining From Images: Constructing a Synthetic Dataset for Multimodal Learning \\ 
\vspace{0.5em} \Large ILSP DCVLR Competition Submission}

\author{
  \textbf{Eleni E. Kasoura, Dimitrios Damianos,} \\
  \textbf{Georgios Paraskevopoulos and Vassilios Katsouros} \thanks{Use footnote for providing further information...} \\
  \vspace{0.2cm} \\
  Institute of Language and Speech Processing \\
  Athena Research Center \\
  Athens, GR 151 25 \\
  \texttt{\{eleni.kasoura, d.damianos, g.paraskevopoulos, vsk\}@athenarc.gr}
}

\begin{document}

\maketitle

\begin{abstract}
This work presents a compact and reproducible pipeline for synthesizing and curating image-based instruction-response pairs for DCVLR. We draw from the Cauldron collection and use Magpie-style instruction and response synthesis with Qwen2.5-VL-7B-Instruct to create concise, image-conditioned instruction pairs. The pipeline consists of instruction synthesis, conditioned on input images and controlled via a system prompt, sanitization, tagging, and filtering of the instructions based on the score obtained by a reward model (DeBERTa-v3). We retain samples with positive scores to maximize the quality of the generated instructions. We observe that score distributions have a median and mean reward score of $\sim1.53$ and $\sim1.75$ respectively, indicating that most generated instructions are of good quality. The resulting dataset contains diverse instructions in the math, spatial, and social reasoning domains.
\end{abstract}

\section{Introduction}

Multimodal systems increasingly rely on image-grounded instruction–response data to elicit robust visual reasoning. However, existing resources often prioritize scale over \textit{coverage, cleanliness, and reproducibility}, which are equally critical for the DCVLR competition. We ask: \emph{Can we build a compact, high-quality, and fully traceable image–instruction corpus that reliably triggers visual reasoning without sacrificing efficiency?}

Here, we present a single-pass, compute-efficient pipeline that synthesizes concise, vision-aware instructions and high-fidelity responses with explicit traceability and tunable quality control.

Our approach adapts recent synthesis pipelines (i.e., Magpie~\cite{xumagpie}) to vision–language models (VLMs). Starting from the \texttt{Cauldron} collection~\cite{laurenccon2024matters}, we condition the Magpie synthesis pipeline on the images to generate instructions and corresponding answers using \texttt{Qwen2.5-VL-7B-Instruct}~\cite{bai2025qwen2} under input formatting and sanitization. The instructions are then tagged, and scored using a reward model to keep only instructions with high reward scores, as a quality control step.

Concretely, the pipeline proceeds through five stages: (1) loading and optional subsampling; (2) instruction synthesis; (3) response generation with vision-aware decoding; (4) instruction sanitization; and (5) reward-model scoring. For multi-image conditioning, we introduce a Magpie-inspired formatting strategy that packs vision tokens to yield concise, grounded interactions without bloated context.

\textbf{Contributions.}

\begin{itemize}
    \item \textbf{Magpie-style pipeline:} We use a system prompt and an image-only pre-query; the model first synthesizes the user instruction from the image, then generates the paired assistant response—yielding grounded instruction–response pairs.
    \item \textbf{Vision-aware formatting for multi-image inputs.} A Magpie-inspired scheme for conditioning on multiple images (via vision tokens) that preserves grounding while keeping interactions concise.
    \item \textbf{Automated, tunable quality control.} A reward-model scoring stage with a tunable threshold $t$. Stricter thresholds lead to higher rejection rates, but increase the final instruction quality
\end{itemize}

 \section{Related Work}

\textbf{Template-Driven Instruction Synthesis.} Aligned language models can \emph{self-synthesize} instructions when prompted with the left context of a chat template, eliminating seed questions and heavy prompt engineering. \textbf{Magpie}~\cite{xumagpie} formalizes this idea and scales instruction creation with simple formatting and cleaning. Earlier works, i.e., \textbf{Self-Instruct}~\cite{wang2023self}, \textbf{Alpaca}~\cite{dubois2023alpacafarm}, and \textbf{Evol-Instruct/WizardLM}~\cite{xu2024wizardlm} have more limited scalability and diversity, regarding the number of synthesized instructions. We adapt Magpie’s self-synthesis to the VLM setting while enforcing strict formatting and sanitization. 
 
\textbf{Vision–Language Instruction Tuning and Formatting.} Modern VLMs adopt concise, chat-oriented interfaces and lightweight post-training. Modern VLM examples tuned on visual instructions include \texttt{LLaVA}~\cite{liu2023visual}, \texttt{InstructBLIP}~\cite{dai2023instructblip}, \texttt{MiniGPT-4}~\cite{zhuminigpt}.
These systems exhibit visual reasoning capabilities and excellent multi-task performance, highlighting the benefits of multi-image conditioning, text-rich inputs, and instruction diversity.

\textbf{Reward Models and Post-Hoc Filtering.} Quality control in alignment often relies on reward models to filter or weigh data before training~\cite{ouyang2022training,bai2022training}. We use a reward model based on \texttt{DeBERTa-v3}~\cite{hedebertav3} for instruction scoring which provides good performance and low latency. Thresholding the scores yields a simple, reproducible keep/cut rule.

 \section{Methodology}
\begin{figure}
    \centering
    \includegraphics[width=0.6\linewidth]{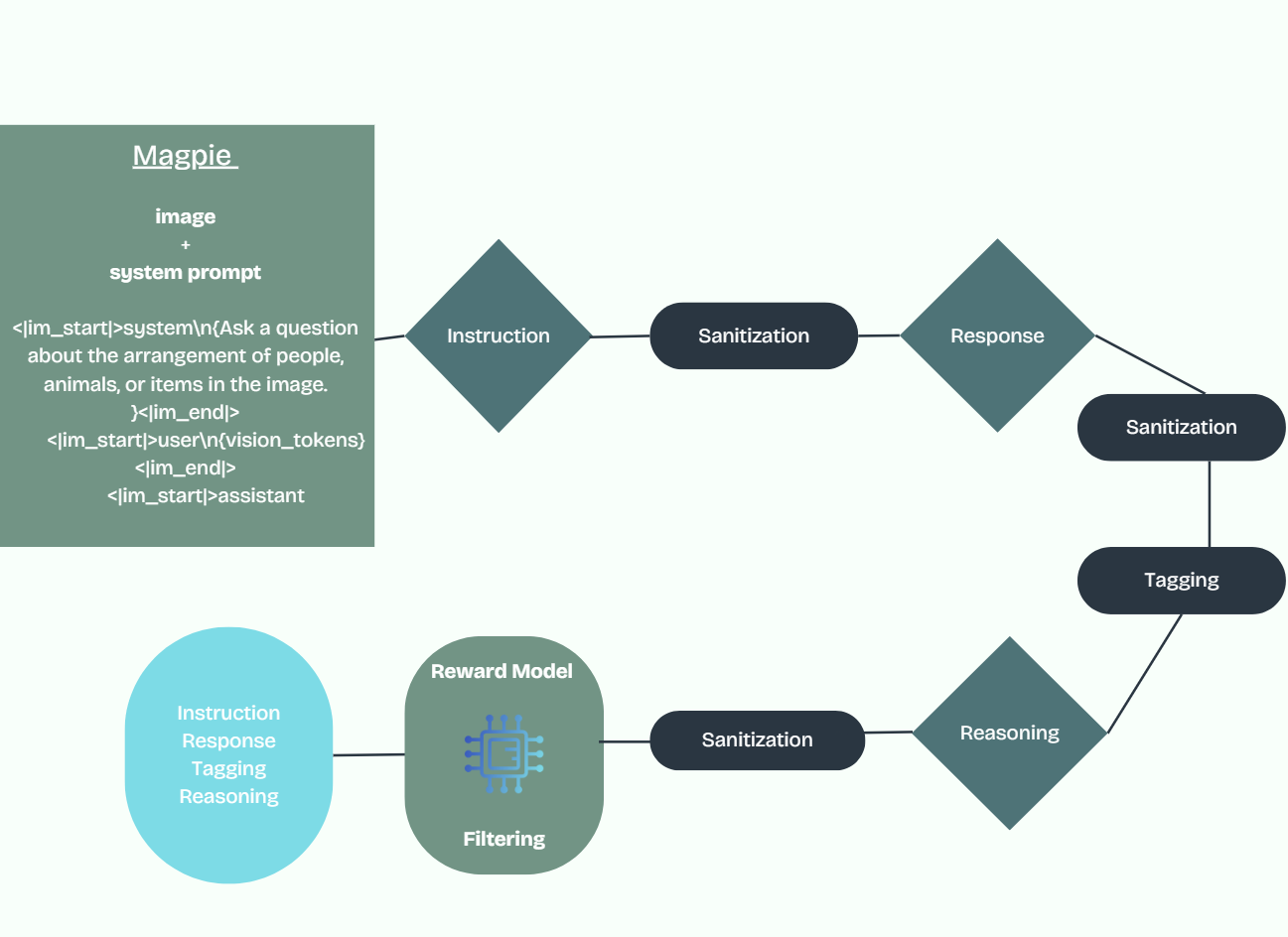}
    \caption{The proposed pipeline for data synthesis and curation based on Magpie.}
    \label{fig:pipeline}
\end{figure}

Fig.~\ref{fig:pipeline} shows the proposed multimodal instruction synthesis pipeline.
We adopt a Magpie-style pipeline that turns images into concise, grounded prompt-response-image triplets, with lightweight cleaning and a reward-based filter.  Concretely, we use an image-only pre-query; the model first autoregressively synthesizes the user instruction based on the image, and then, after inserting the assistant header, generates the paired assistant response. We apply light sanitization (token/whitespace normalization, template validation), rule-based guards for modality grounding, i.e., the instruction must reference visible content, and near-duplicate removal.  
Each example is conditioned by the image and a short system prompt that specifies the task family, e.g., ``ask about spatial arrangement". We use a compact chat format system turn followed by a user turn that carries vision tokens, so the model is primed to emit an instruction directly from visual context (self-synthesis). 

The model produces a candidate instruction conditioned on the image. A sanitization pass removes boilerplate, control markers, and residual formatting, yielding a minimal, model-ready prompt. 
Given the cleaned instruction and the same image context, the model generates a response. A second sanitizationr pass standardizes punctuation and trims generation artifacts. We then perform tagging, e.g., difficulty labels, for downstream analysis. 

All instruction–response pairs pass through a reward model that assigns a scalar score. A simple threshold selects retained items (higher scores) and discards low-quality pairs, enabling stricter or looser quality control. An optional reasoning check can be applied before final acceptance, followed by a brief final sanitization to ensure formatting consistency. 

The outputs of this pipeline are compact, cleaned instructions, responses, and tags, together with reward scores. These artifacts are ready for post-hoc filtering, analysis, and integration into instruction-tuning or evaluation workflows.

\section{Implementation details}

For instruction and response synthesis we use \texttt{Qwen2.5-VL-7B-Instruct}. Images are sampled from the Cauldron collection, a unified collection of $\sim50$ vision–language training sets released with \texttt{Idefics2}~\cite{laurenccon2024matters}. Cauldron aggregates multi-domain tasks (VQA, OCR/docs, charts/tables, math/logic, multi-image, screenshots) and is widely reused for instruction tuning. We sample $5000$ images from each subset and generate instructions based on the sampled images. The reward model used for scoring is \texttt{OpenAssistant/reward-model-deberta-v3-large-v2}\footnote{\url{https://huggingface.co/OpenAssistant/reward-model-deberta-v3-large-v2}}.
We set the selection threshold to $t=0$, keeping samples with positive scores, and then select the top $10000$ samples among those. All experiments are run on nodes from the Leonardo-Booster supercomputer. Each node contains $4$ NVIDIA A100 GPUs with 64GB VRAM.





\subsection{System Prompts}

System prompts are organized into categories to steer generation for different capabilities. We maintain targeted categories for \emph{text extraction}, \emph{spatial relations}, and \emph{mathematics/quantitative} reasoning, alongside a \emph{general} category that applies across most subsets as seen in Appendix~\ref{sec:appendix}.


\begin{figure}[h]
    \centering
    \includegraphics[width=0.55\textwidth]{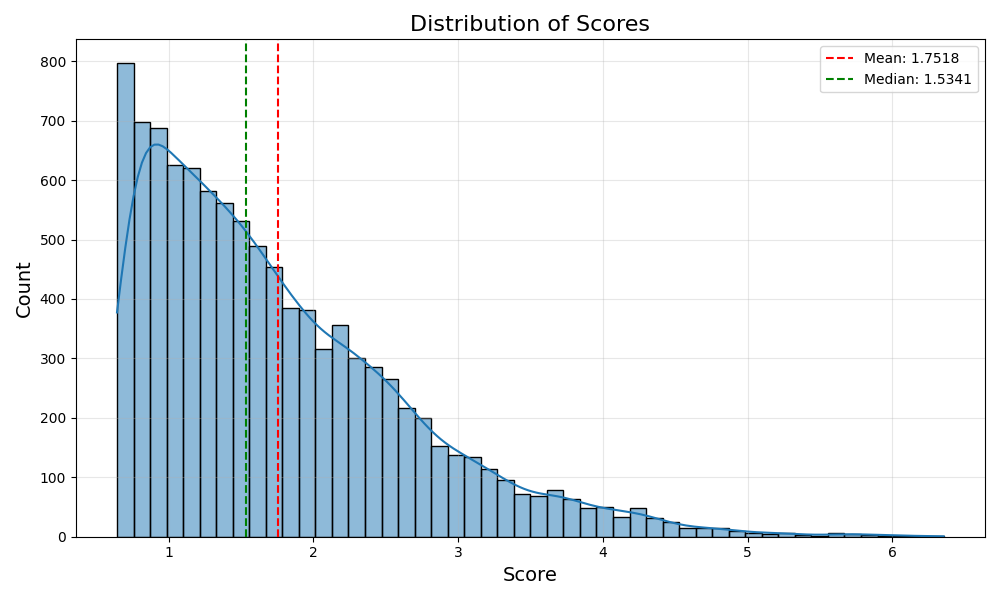}
  \caption{Reward score distribution of selected samples}
  \label{fig:scores}
\end{figure}

\begin{table}[ht]
  \centering
  \caption{Subset counts for sampes included in the generated dataset. The dataset includes 10000 samples in total.}
  \label{tab:subset_counts}
  \begin{tabular}{lr}
    \toprule
    Subset & Count \\
    \midrule
    cocoqa      & 1354 \\
    aokvqa      & 1311 \\
    vqav2       & 1289 \\
    clevr       & 998 \\
    ocrvqa      & 971 \\
    ai2d        & 695 \\
    vsr         & 539 \\
    robut       & 516 \\
    mapqa       & 396 \\
    infographic & 377 \\
    vistext     & 260 \\
    chart2text  & 198 \\
    figureqa    & 197 \\
    iconqa      & 192 \\
    chartqa     & 185 \\
    vqarad      & 103 \\
    diagram\_image\_to\_text & 82 \\
    tat         & 81 \\
    hitab       & 73 \\
    hateful     & 70 \\
    finqa       & 61 \\
    dvqa        & 52 \\
    \midrule
    \textbf{Total} & \textbf{10\,000} \\
    \bottomrule
  \end{tabular}
\end{table}


\section{Dataset Statistics}




In Fig.~\ref{fig:scores} we see the reward score distribution of the selected samples. We see that the average and median reward scores are $1.7518$ and $1.5341$ respectively. with sample scores ranging from $\sim0.5$ to $\sim6.5$.
Table~\ref{tab:subset_counts} presents the subset counts for samples included in the final dataset. We observe that the dataset is skewed towards Visual Question Answering samples, while subsets containing math, diagrams and infographics are less represented. One explanation for this is that due to the higher complexity of these images, the model generates lower quality instructions and they are rejected by the reward model.

\section{Conclusions}
In this work we extend Magpie to generate visual instructions grounded on input images. Instruction synthesis is guided by system prompts and the quality of the resulting samples is automatically determined through reward model scores. In our implementation we have observed that proper sanitization of the generated instructions and the inclusion of system prompts for conditioning is paramount, since VLMs are biased towards producing captions and descriptions of the input images without further guidance.

In the future we plan to extend this pipeline to generate more diverse visual instruction sets, through the inclusion of more system prompts and diverse input images. Furthermore, we plan to run detailed ablations to guide our decisions on reward model, threshold and system prompt selection.


\section*{Acknowledgements}
We acknowledge the EuroHPC Joint Undertaking for awarding this project access to the EuroHPC supercomputer LEONARDO, hosted by CINECA (Italy) and the LEONARDO
consortium through an EuroHPC Development Access call.

\section*{Limitations}

Our work is still at the proof of concept stage and has several limitations. First, due to time constraints ablation studies were not systematically performed. The model used for synthesis is a relatively small (7 billion) parameter LLM, limiting the quality and diversity of the synthesized visual instructions. Finally, stratification of the selected samples during reward scoring and inclusion of more system prompts may be included to ensure higher diversity of the final dataset.

\section*{Broader Impact}

Our work on synthetic data generation has broad societal implications, offering substantial benefits while also posing serious risks. On the positive side, this technology can significantly accelerate machine learning research, since scalable data synthesis pipelines are an economically viable way to obtain data, especially for areas and problems where data are relatively scarce. Furthermore, synthetic data offer pathways for more private and fair models, since inherent biases can be balanced out and personal identifying information that exists in real datasets is not included. However, these capabilities are shadowed by the potential for malicious use, particularly in the generation of sophisticated disinformation and deepfakes. Furthermore, improperly trained models risk amplifying the societal biases they are intended to correct. Progress in this domain must be tightly coupled with a commitment to developing detection techniques, fairness audits, and clear ethical guidelines. Finally, there is an ongoing discussion about the decline in diversity of LLM outputs due the self-consuming training loops in the research community, and synthetic data contribute to this issue. Further investigation should be performed about the limits of this approach.

\bibliographystyle{plainnat}
\bibliography{refs}

\appendix
\section{System prompt categories}
\label{sec:appendix}


\begin{table}[ht]
\centering
\begin{tabular}{p{0.23\textwidth} | p{0.72\textwidth}}
\toprule
\textbf{Prompt type} & \textbf{Prompt examples (2 per type)} \\
\midrule
prompts &
  Try to ask an easy question for this image. \\
  &Create an instruction that prompts the user to explain the sequence of events taking place in the image.\\
  \hline
math\_prompts &
  Create a question that asks the user to describe the mathematical concept represented in the diagram. \\
  & Write an instruction that asks the user to summarize the main statistical findings illustrated by this chart.\\
\hline
multi\_image &
   Compare the 2 images and write a question that highlights the differences in the objects or scenes depicted. \\
  & Look at the 2 images and create an instruction that asks the user to identify the similarities between the two. \\
\hline
general\_prompts &
  Ask a question about one of the objects in this image.\\
 & Create a question about what the people or animals are doing in this picture.\\
\hline
spatial\_prompts &
   Write a question that asks the user to describe where the main objects are located in relation to each other. \\
  & Generate an instruction that asks the user to explain whether this is an indoor or outdoor scene.
\\
\hline
text\_prompts &
   Generate a question that asks the user to read and summarize the visible text in the image. \\
  & Write an instruction that asks the user to transcribe the letters or words shown in this picture.
\\
\bottomrule
\end{tabular}
\caption{Prompt categories with two example prompts per type.}
\label{tab:prompt_types_examples}
\end{table}

\end{document}